# Evolving Genetic Programming Tree Models for Predicting the Mechanical Properties of Green Fibers for Better Biocomposite Materials

Faris M. AL-Oqla[1], Hossam Faris [2,3], Maria Habib[4], Pedro A. Castillo-Valdivieso[5]

[1]Department of Mechanical Engineering, Faculty of Engineering, The Hashemite University, P.O box 330127, Zarqa 13133, Jordan,

Fmaloqla@hu.edu.jo

[2]King Abdullah II School for Information Technology, The University of Jordan, Amman, Jordan

[3]Research Centre for Information and Communications Technologies of the University of Granada

(CITIC-UGR), University of Granada, 18011 Granada, Spain

hossam.faris@ju.edu.jo

[4]Doctoral School of Sciences, Technologies, and Engineering, the University of Granada, 18011

Granada, Spain, mhabib@correo.ugr.es

[5]Department of Computer Engineering, Automatics and Robotics, ETSIIT-CITIC, University of Granada, 18011 Granada, Spain

**Abstract**

Advanced modern technology and industrial sustainability theme have contributed implementing composite materials for various industrial applications. Green composites are among the desired alternatives for the green products. However, to properly control the performance of the green composites, predicting their constituents properties are of paramount importance. This work presents an innovative evolving genetic programming tree models for predicting the mechanical properties of natural fibers based upon several inherent chemical and physical properties. Cellulose, hemicellulose, lignin and moisture contents as well as the Microfibrillar angle of various natural fibers were considered to establish the prediction models. A one-hold-out methodology was applied for training/testing phases. Robust models were developed to predict the tensile strength, Young's modulus, and the elongation at break properties of the natural fibers. It was revealed that Microfibrillar angle was dominant and capable of determining the ultimate tensile strength of the natural fibers by 44.7% comparable to other considered properties, while the impact of cellulose content in the model was only 35.6%. This in order would facilitate utilizing artificial intelligence in predicting the overall mechanical properties of natural fibers without experimental efforts and cost to enhance developing better green composite materials for various industrial applications.

# 1. Introduction

Modern green products and sustainable industry have put out the most effort in the development and use of composite technology. The requirement for robust, stronger, and lighter constructions offered a chance for composite materials to outperform other regularly utilized materials[1-3]. Manufacturing and maintenance expenses have been reduced because of the development of new and improved fabrication techniques in this sector. In addition, composites have had the largest influence on the athletic equipment and sustainability[4-6]. Rapidly displacing traditional materials is made possible by green products using natural fiber reinforced polymeric composite materials; for example, automotive applications, packaging applications, and a variety of sport goods are now virtually entirely made of innovative composites[7-12]. Without question, the utilization of green composite materials will increase in the future as a requirement of the modern societies as well as the environmental issues. However, the characteristics of the green product strongly depend upon the constituents' individual characteristics as well as their compatibility.

The chemical composition of the fibers controls their interactions with the polymer matrices, and substantially, highly affects the composite properties. Chemical composition of commonly used natural fibers and their mechanical and physical properties are listed in Table 2 [13-22].

Table 1: Various properties of natural fibers

|  | Cellulose (wt.%) | Hemicelluloses (wt.%) | Lignin (wt.%) | Moisture Content (wt.%) | Microfibrillar Angle (deg) | Young's modulus (GPa) | Ultimate tensile strength (MPa) | Elongation at break (%) |
|---|---|---|---|---|---|---|---|---|
| **Flax** | 71 | 18.6-20.6 | 2.2 | 8-12 | 5-10 | 27.6 | 345-1500 | 2.7-3.2 |
| **Hemp** | 70-74 | 17.9-22.4 | 3.7-5.7 | 6.2-12 | 2-6.2 | 70 | 690 | 1.6 |
| **Jute** | 61-71.5 | 13.6-20.4 | 12- 13 | 12.5-13.7 | 8 | 13-26.5 | 393-800 | 1.16-1.5 |
| **Kenaf** | 45-57 | 21.5 | 8 - 13 | 6.2-9.1 | 7.1-15 | 53 | 930 | 1.6 |
| **Ramie** | 68.6-76.2 | 13.1-16.7 | 0.6-0.7 | 7.5-17 | 7.5 | 61.4-128 | 400-938 | 1.2-3.8 |
| **Sisal** | 66-78 | 10 - 14 | 10 - 14 | 10 - 22 | 10 .22 | 9.4-22 | 468-700 | 3.0-7.0 |
| **Banana** | 63-64 | 10 | 5 | 10 - 12 | 10-12 | 12- 33 | 355-500 | 1.5-9 |
| **Oil palm** | 50-65 | 30 | 17-19 | 11-29 | 42-46 | 3.2 | 248 | 2.5 |
| **Cotton** | 82.7-95 | 2-15 | 0.1-2 | 7.85-8.5 | 33-34 | 287-800 | 5.5-12.6 | 3-10 |
| **Coir** | 32-45 | 0.15-0.25 | 40-45 | 8 | 30-49 | 131-220 | 4-6 | 15-40 |

Cellulose, hemi-cellulose, lignin, and moisture are the main chemical composition of the natural fibers. However, with different ratios, even for the same plant at the various positions[8; 23-25]. These ratios as well as the Microfibrillar angle highly influence the fiber mechanical performance and thus, affect the overall green composites properties. As the strength of the fiber is the most important feature of the natural fiber to be considered in green composites, the selection of the fibers is vital in such type of materials. In order to property select the natural fiber type, proper prediction of its mechanical performance is important [26-30].

The resolution of this type of computational prediction problems can be approached by using various advanced computational methods. These include genetic programming (GP). GP is an evolutionary algorithm that is inspired by the Darwinian principles of evolution and natural selection. Originally, it was proposed as an extension of the genetic algorithm (GA). Primarily, the objective was to automatically solve problems by initially knowing abstract general information of the solutions. Since its proposal, it has been applied to various applications including image processing, industrial modeling and control, finance, and bioinformatics [31-35]. However, it has a special success and a wide-adoption in fitting regression models, where often there is a weak understanding of the relationships among the features of the respective problem [36; 37]. The GP evolutionary algorithm shares many similarities with the GA algorithm, which consists of a population of solutions that evolve iteratively by genetic operators. Meanwhile, the critical difference is the representation of the individuals (solutions) as tree structures and not strings of bits. Unlike the GA algorithm that uses fixed-length individuals, the GP algorithm represents the individuals using hierarchical and variable-length solutions that are more capable of modeling the tasks of a computer program. The GP algorithm creates a population of computer programs (tree-based individuals) that evolve over generations using genetic operators (i.e., selection, subtree crossover, and subtree mutation) that make is suitable for various engineering applications.

Consequently, this work introduces a novel evolving GP tree models for predicting the mechanical properties of green fibers based upon several intrinsic chemical and physical properties. Here, several chemical compositions of cellulose, hemicellulose, lignin and moisture content as well as the Microfibrillar angle of various natural fibers are considered to be investigated and building the prediction models. A one-hold-out methodology is applied for training/testing in the developed prediction models. Moreover, in order to obtain more reliable results, the one-hold-out is repeated 10 times to enhance the reliability of the established models of the impact of the fibers chemical and physical properties on the tensile strength, modulus of elasticity, and the elongation at break properties. The impact of each chemical property parameter of natural fibers in predicting their mechanical performance was determined using the genetic programing as one of the artificial intelligence methods in this field. This in order would facilitate predicting the overall mechanical properties of natural fibers without both experimental efforts and cost to enhance their proper selection for the green composite materials to develop more sustainable green products.

## 2. Materials and Methods

### 2.1. Data collection

In order to establish proper evolving GP tree models for predicting the mechanical characteristics of green fibers, several intrinsic chemical and physical properties of the cellulosic fibers were collected from reliable experimental works found in the literature. This includes cellulose, hemicellulose, lignin, moisture content and Microfibrillar angle of the natural fibers worldwide. To enhance the reliability of the work, data was only collected from peer reviewed journals indexed in Scopus and Clarivate Analytics. It is a worthy notice that there was a variety in the reported values for almost all the considered fibers due to the variations of the fiber type, age, place, climate, soil and fertilizers. This in fact demonstrates the complexity of performing such experimental works for most of the available natural fibers as well as revealing the importance of establishing prediction models to determine the most important factors that affects the overall mechanical performance of such fibers.

### 2.2. GP procedure

Conceptually, the GP algorithm shares the same procedural structure of other algorithms in the field of evolutionary computation. Initially, any evolutionary algorithm has a population of individuals (i.e., solutions) that evolve over generations to converge for optimal regions and find the best solutions. The evolutionary processes involve the probabilistic selection of the parent solutions, the reproduction (crossover), the mutation, and the elitism. However, the GP algorithm has a population of computer programs, which are represented by a tree structure, and then evolve using subtree crossover and subtree mutation operations. The following steps discuss more closely the GP algorithm and its genetic operators. Figure 1 illustrates the abstract flow of the GP algorithm.

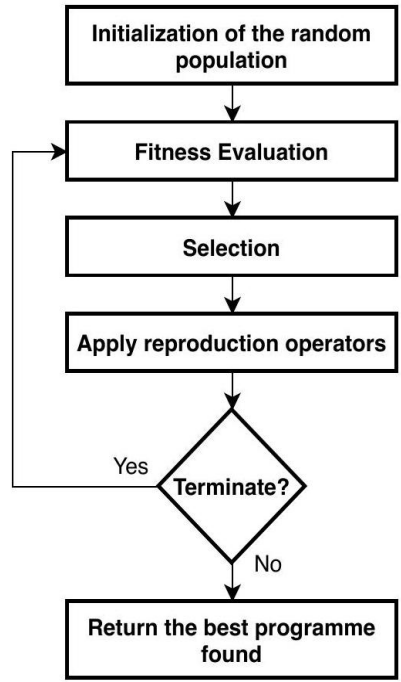

Figure 1: A flowchart of the GP algorithm

- Initialization: this includes the initial creation of a population of computer programmes or individuals. Since those individuals are modeled in a tree structure; the internal nodes of the tree are represented by a set of function symbols, while the leaves are represented by a set of terminal symbols. The function symbols are mathematical functions (e.g., arithmetic operations, conditional functions, or sinusoidal functions). Whereas, the terminal symbols are variables or constant values. In Figure 2, subfigure (a) shows an example of an individual of the GP algorithm that is represented by the following mathematical model: $(1.5 - (Y + 2)) + (X/3)$. However, different methods were proposed for the initialization of the tree individuals examples of which include the ramped half-and-half and the grow methods.

- Selection: this operation decides which individuals will be transferred for the next iteration to generate a new population of individuals. Broadly, the selection process might be random-based (e.g., tournament selection) or ranking-based depending on the fitness scores of individuals. However, this step is critical, since it influences the diversity of the next generation, which considerably affects the overall performance of the algorithm.

- Crossover: uses the selected parents for producing new variants of individuals by exchanging their genetic material. The GP crossover is denoted by the subtree crossover, which has a randomly selected crossover point (node) at each parent individual. Although crossover points are selected randomly, they are preferred to be far from the root or the leaves to avoid increasing the complexity of the new offspring. In the one-point crossover, the parents at the crossover points swap their subtrees to produce new individuals as illustrated in Figure 2 (c).

- Mutation (subtree mutation): this acts only on a single parent individual, where the selected subtree at a randomly selected mutation point is replaced by another random subtree. This is demonstrated by subfigures (a) and (b) in Figure 2, where (b) is a new mutated parent individual.

- Elitism: this operation preserves the best-found individual(s) at each generation.

- Fitness assessment (Objective Function): Every solution (individual or tree) of the population is assessed and assigned a score that indicates its capability to address the targeted problem. This score is known by the fitness value of an individual and refers to the optimization objective of the problem. In other words, in prediction problems where the aim is to minimize the error; the mean absolute error (MAE), the root mean square error(MSE), or the ratio of error can be used as the objective method.

- Termination: the evolutionary process is iterative. The algorithm loops over the evolutionary operations from selection, to reproduction (crossover and mutation), and then fitness evaluation over the course of generations until an optimal solution is reached or a termination condition is satisfied. Often, the maximum number of iterations is used as a criterion for stopping the optimization procedure.

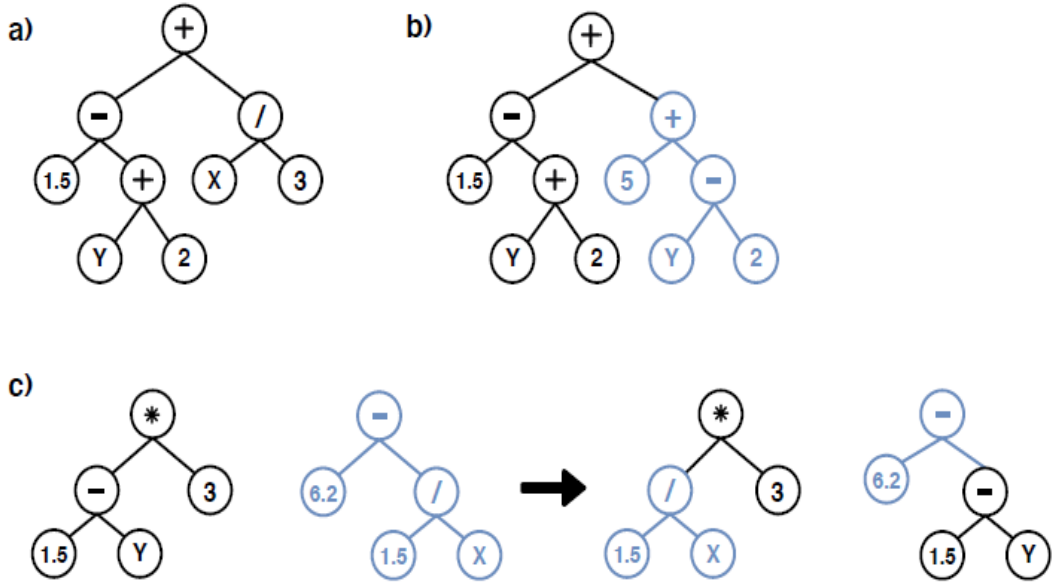

Figure 2: A representation of an individual of GP algorithm, in addition to the subtree mutation and subtree crossover.

### 2.3. Identification of the most relevant features

A main advantage of GP is its availability to evolve tree based model that are easy to interpret. By nature, GP can provide a better understanding of the most significant variables involved in the prediction process through an incorporated feature selection mechanism. Over the course of the evolutionary cycle of GP, some variables will survive and have higher probability to appear in later generations, while variables with less impact will gradually disappear. To determine the most important features in the developed GP tree models, the approach will be referred to as the measurement of "relative impact" of input variables [38; 39]. The relative impact on an input variable is measured according to the number of references to this variable in all generated GP models starting from the initial population until the last generation. The relative impact of a given input variable can be measured as follows.

- Let $S$ be a set of input variables $\{x_1, x_2, ..., x_n\}$ where $x_i \in S$.

- Through the iteration of GP, the number of appearannces of of $x_i$ concerning a model $M$ can be denoted as RefCount $(x_i, M)$.

- The occurrence rate of $x_i$ in population $P$ can be formulated as:

$$\text{freq}(x_i, P) = \sum_{M \in P} \text{Ref Count}(x_i, M) \qquad (1)$$

- As a result, Equation 2 shows that the proportion of references of variable $x_i$ in a population $P$, represented by freq $(x_i, P)$, is divided by the total number of variable references to determine the relative frequency, rel $(x_i, P)$, of variable $x_i$ in $P$

$$\text{rel } (x_i, P) = \frac{\text{freq } (x_i, P)}{\sum_{k=1}^{n} \text{freq } (x_k, P)} \tag{2}$$

## 2.4. Evaluation metrics

The evaluation metrics utilized in this study to evaluate the ultimately attained GP models are:

1. Root Mean Squared Error (RMSE):

$$RMSE = \sqrt{\frac{1}{n}\sum_{i=1}^{n} (y_i - \hat{y}_i)^2} \tag{3}$$

2. Mean Absolute Error (MAE):

$$MAE = \frac{1}{n}\sum_{i=1}^{n} |y_i - \hat{y}_i| \tag{4}$$

3. Coefficient of Determination $(R^2)$:

$$R^2 = \frac{\sum_{i=1}^{n} (\hat{y}_i - \overline{y_i})^2}{\sqrt{\sum_{i=1}^{n} (\hat{y}_i - \overline{y_i})^2 + \sum_{i=1}^{n} (\hat{y}_i - y_i)^2}} \tag{5}$$

where $y$ and $\hat{y}$ are the actual and the predicted values based on the developed GP models, and $n$ is the number of instances used in the experiments.

## 3. Results and Discussion

In order to generate prediction models for the mechanical performance of green fibers based upon the intrinsic fiber characteristics using GP method, a one-hold-out methodology is applied for training/testing. That is, all instances are used for training except one instance left for testing. This process is repeated ten times, each time with different training and testing instances to develop more reliable results.

### 3.1. Ultimate tensile strength GP model

In order to demonstrate the impact of the intrinsic features of the green fiber on the ultimate tensile strength mechanical property, GP prediction model was established considering all of cellulose (C), hemicellulose (H), lignin (L), moisture content (Mc), and Microfibrillar angle (Ma) of the fiber simultaneously. The best generated GP model is expressed by the following equation.

$$UTS = \left((\log(c_0 \cdot H) \cdot (c_1 \cdot Ma - c_2 \cdot C) - c_3 \cdot Ma) \cdot c_4 + c_5\right) \quad (6)$$

$$
\begin{aligned}
c_0 &= 1.378 \\
c_1 &= 0.94995 \\
c_2 &= 0.78676 \\
c_3 &= 1.8922 \\
c_4 &= -6.3671 \\
c_5 &= -363.56
\end{aligned}
$$

It can be demonstrated from the best GP prediction model that the ultimate tensile strength was mainly influenced by the cellulose content, Microfibrillar angle and to some extent by the hemicellulose of the fiber, but not influenced by moisture content and lignin. The average and best GP results for the ultimate tensile strength case in terms of RMSE, MAE and $R^2$ are given in Table 2. The estimated values for ultimate tensile strength by the developed GP models of the best experiment are given in Table 3. The actual vs. estimated ultimate tensile strength values by best one-hold-out cross-validation GP experiment are expressed in Figure 3. Moreover, the relative impact for each variable identified by GP over the course of iteration is shown in Figure 4. It can also be confirmed that both cellulose content and the Microfibrillar angle of the fiber have the main influence in determining the ultimate tensile strength of the natural fibers. The Microfibrillar angle was the dominant in determining the ultimate tensile strength of the natural fibers by 44.7%, and cellulose content was the second by 35.6%. However, all of hemicellulose, moisture content and lignin have very minor effect in this mechanical property of the fibers.

Table 2: Evaluation results of GP and MLR for modeling the ultimate tensile strength.

|       | LR     | Best GP | Average GP |
|-------|--------|---------|------------|
| RMSE  | 385.55 | 129.99  | 196.6      |
| MAE   | 308.66 | 97.38   | 138.3      |
| $R^2$ | 0.0003 | 0.65    | 0.3        |

Table 3: Estimated values by the best one-hold-out cross-validation experiment for ultimate tensile strength.

| Ranges     | Actual | Estimated |
|------------|--------|-----------|
| 345 − 1500 | 922.5  | 628.1     |
| 690 | 690 | 725.1 |
| 393 − 800 | 596.5 | 670.2 |
| 400 − 938 | 669 | 675.7 |
| 468 − 700 | 584 | 555.0 |
| 355 − 500 | 427.5 | 622.9 |
| 248 | 248 | 337.9 |
| 287 − 800 | 543.5 | 467.3 |
| 131 − 220 | 175.5 | 99.4 |

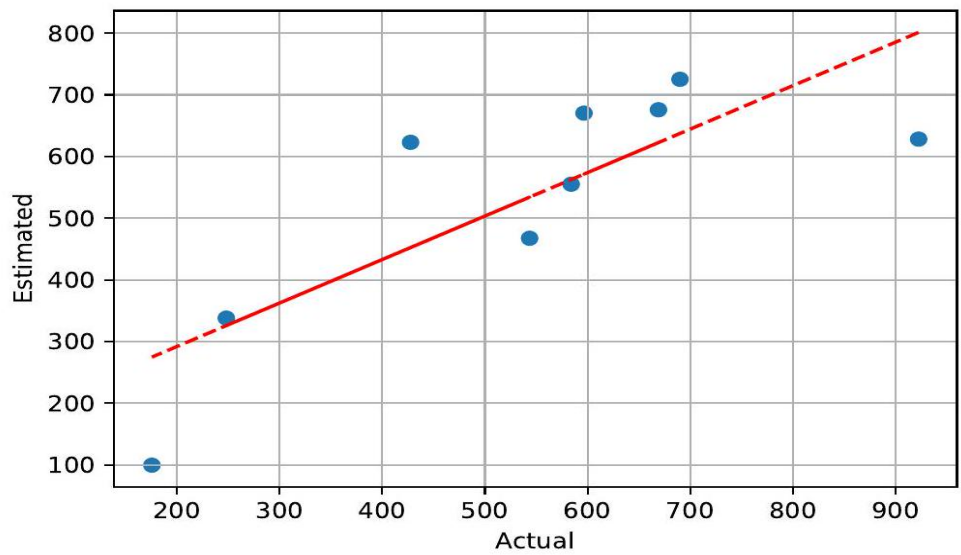

Figure 3: Actual vs. estimated Ultimate tensile strength values by best one-hold-out cross-validation GP experiment.

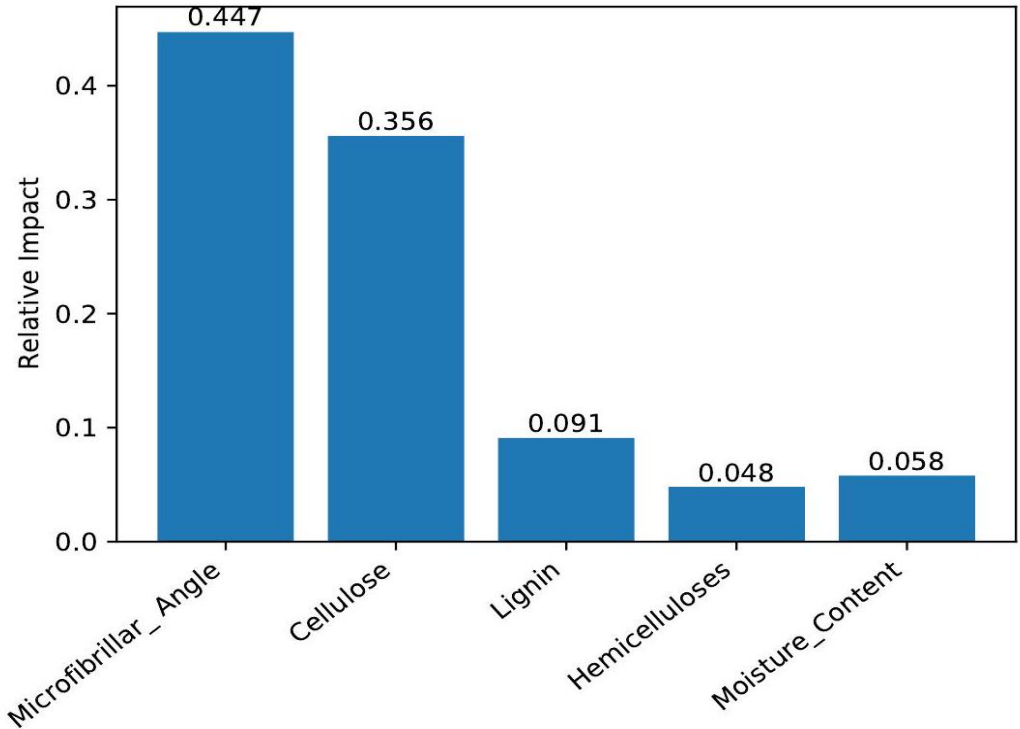

Figure 4: Relative impact of the mechanical properties identified by GP for the ultimate tensile strength model.

## 3.2. Elongation at break (%)

The elongation at break property of the natural fibers was also investigated via GP models. The actual vs. predicted elongation at break values by best one-hold-out cross validation GP experiment is illustrated in Figure 5. It can be shown that despite of existence of extreme values, the predicted values were with high accuracy utilizing the evolved GP tree model. Table 4 demonstrated the evaluation results of GP and MLR for modeling the elongation at break (%). The R2 values for the Best GP and the Average GP were 0.968 and 0.855 respectively. This indicates that the prediction modes were capable of predicting the elongation at break property values on the natural fibers with high confidence comparable to the linear regression model.

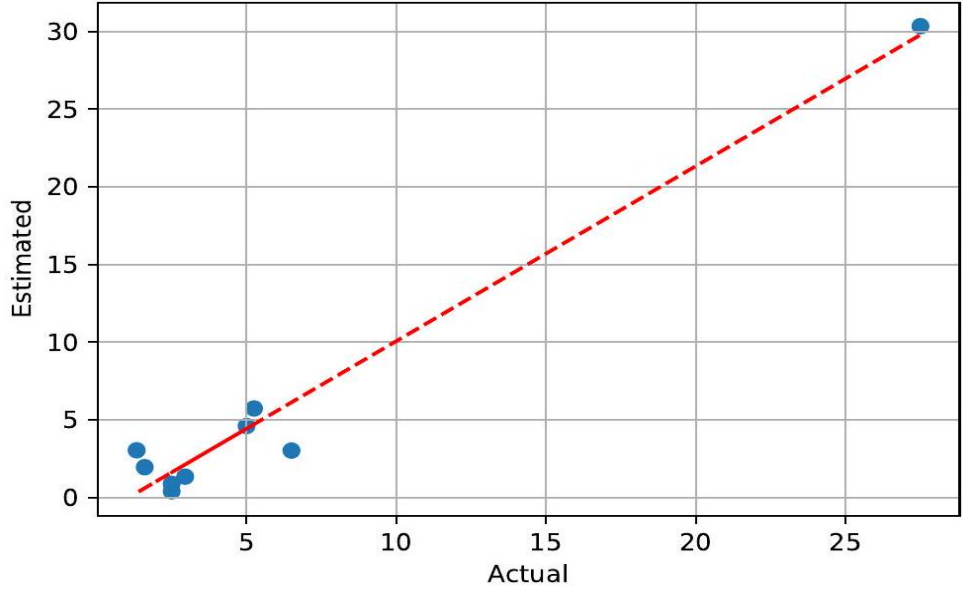

Figure 5: Actual vs. estimated Elongation at break values by best one-hold-out cross validation GP experiment.

Table 4: Evaluation results of GP and MLR for modeling the Elongation at break (%).

|      | LR     | Best GP | Average GP |
|------|--------|---------|------------|
| RMSE | 11.71  | 2.16    | 4.0        |
| MAE  | 8.67   | 1.62    | 2.55       |
| R2   | 0.0125 | 0.968   | 0.855      |

Moreover, the GP elongation at break model is expresses as in Equation 7 with it corresponding constant values. It can be shown that the elongation at break property of the natural fibers was modeled by GP to be a function of hemicellulose and moisture content. However, the rest of compositions like cellulose content, Microfibrillar angle and lignin were not influentially dominant in the GP model.

$$\text{Elong.} = \left(\left(c_0 \cdot H + \frac{c_1 \cdot Mc}{c_2 \cdot H}\right) \cdot c_3 + c_4\right) \qquad (7)$$

$$c_0 = 1.1682$$
$$c_1 = 0.19393$$
$$c_2 = -0.024258$$
$$c_3 = -0.073047$$
$$c_4 = 4.3192$$

The relative impact of the mechanical properties identified by GP for the elongation at break (%) model is illustrated in Figure 6. It is shown that moisture content of the natural fiber has the main influence in determining this property relative to other contents with about 63% of the model. Hemicellulose content is also affected the elongation property with about 29.4%. However, cellulose content, Microfibrillar angle and lignin were not found dramatically influencing the elongation at break property. Cellulose for instance, has only 3.4% relative impact in the GP predicting model. The estimated values by the best one-hold-out cross-validation experiment for elongation at break with various ranges of the fiber properties are tabulated in Table 5.

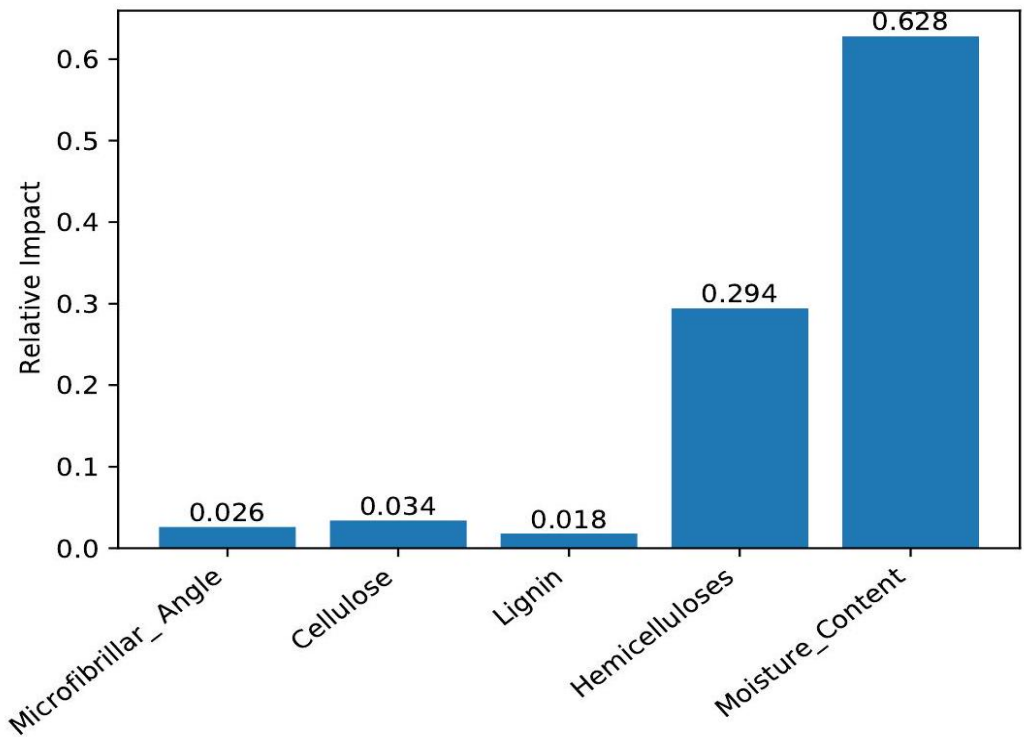

Figure 6: Relative impact of the mechanical properties identified by GP for the elongation at break (%) model.

Table 5: Estimated values by the best one-hold-out cross-validation experiment for Elongation at break.

| Ranges | Actual | Estimated |
|---|---|---|
| 2.7 − 3.2 | 2.95 | 1.34 |
| 1.6 | 1.6 | 1.95 |
| 1.16 − 1.5 | 1.33 | 3.04 |
| 1.2 − 3.8 | 2.5 | 0.40 |
| 3.0 − 7.0 | 5 | 4.60 |
| 1.5 − 9 | 5.25 | 5.73 |
| 2.5 | 2.5 | 0.88 |
| 3 − 10 | 6.5 | 3.02 |
| 15 − 40 | 27.5 | 30.34 |

### 3.3. Young's Modulus

On the other hand, the GP predicted model of the Young's modulus of the natural fibers is expressed as in Equation 8. It can be shown that the best GP model was generated utilizing all of Microfibrillar angle, lignin and hemicellulose.

$$\text{Young 's Modulus} = \left(\frac{\exp(c_0 \cdot \text{Ma})}{(c_1 \cdot \text{L}) \cdot (c_2 \cdot \text{H})} \cdot c_3 + c_4\right) \tag{8}$$

$$\begin{aligned} c_0 &= -0.52777 \\ c_1 &= 2.9897 \\ c_2 &= 3.1136 \\ c_3 &= 3.989E + 05 \\ c_4 &= 12.823 \end{aligned}$$

The estimated values by the best one-hold-out cross-validation experiment for Young's Modulus are tabulated in Table 6 and the actual vs. estimated Young's Modulus values by best one-hold-out cross validation GP experiment are shown in Figure 7. It can be demonstrated that despite of being very difficult to predict the Young's modulus of the natural fibers based on their intrinsic chemical composition, the GP model was capable of doing so with acceptable relative errors. Moreover, the lignin content of the fibers and the hemicellulose were the dominants in the GP predicting model as demonstrated in Figure 8.

Table 6: Estimated values by the best one-hold-out cross-validation experiment for Young's

Modulus.

| Ranges | Actual | Estimated |
| --- | --- | --- |
| 27.6 | 27.6 | 35.43 |
| 70 | 70 | 41.12 |
| 13 − 26.5 | 19.75 | 14.86 |
| 61.4 − 128 | 94.7 | 26.92 |
| 9.4 − 22 | 15.7 | 12.65 |
| 12 − 33 | 22.5 | 12.88 |
| 3.2 | 3.2 | 13.67 |
| 5.5 − 12.6 | 18.1 | 25.69 |
| 4 − 6 | 5 | 21.14 |

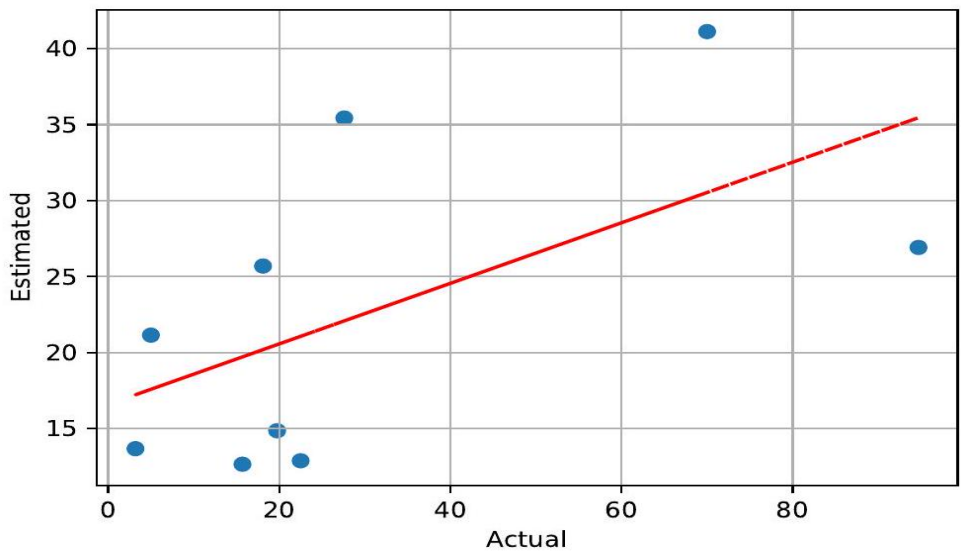

Figure 7: Actual vs. estimated Young's Modulus values by best one-hold-out cross validation GP experiment.

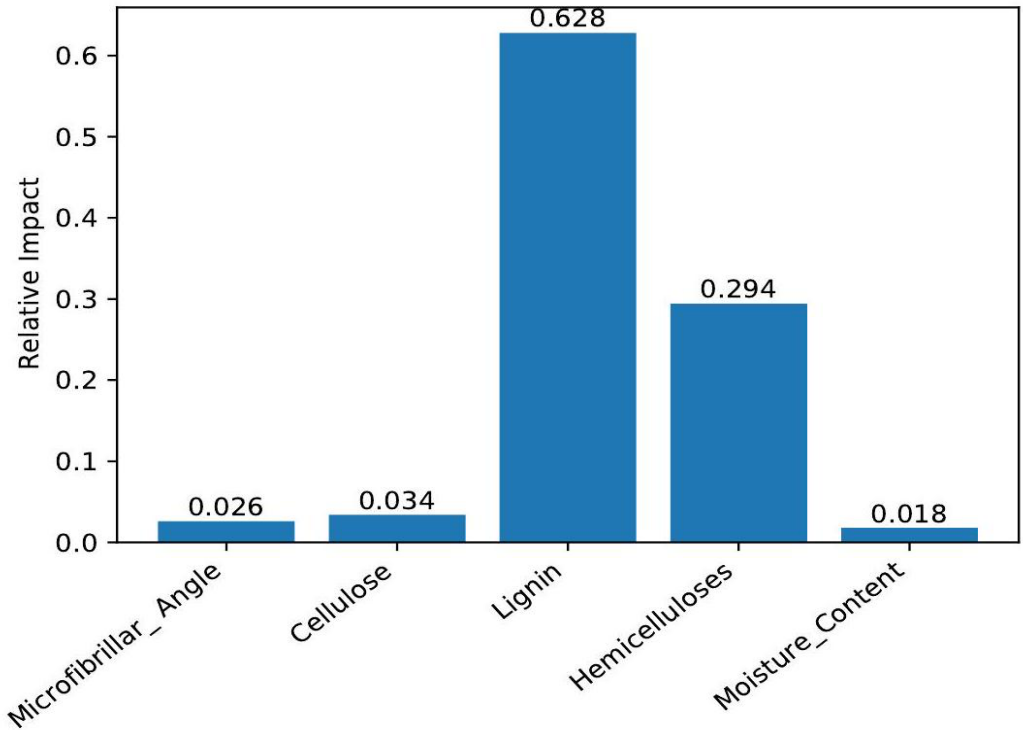

Figure 8: Relative impact of the mechanical properties identified by GP for the Young's modulus model.

## 4. Conclusions

Despite of the complexity in predicting the mechanical performance of green fibers due to their various physical and chemical compositions, this work was capable of establishing evolving GP tree models for expecting the mechanical properties of green fibers. Each mechanical property of the natural fibers was found to be primarily influenced by certain intrinsic properties. It was shown from the best GP prediction models that the ultimate tensile strength was mainly influenced by the cellulose content, Microfibrillar angle and hemicellulose of the fiber, but not influenced by moisture and lignin contents. However, moisture content of the natural fiber has the main influence in determining the elongation at break property relative to other contents with about 63% dominance in the model. Hemicellulose content also affects the elongation at break property with about 29.4%. However, cellulose content, Microfibrillar angle and lignin were not found influencing this property. Moreover, hemicellulose and lignin contents of fibers were found significant in determining the Young's modulus property according to the established GP prediction models. This in order would facilitate proper selection of the natural fibers for desired green composites that fulfil the sustainable industrial requirements and customer satisfaction attributes.

### Acknowledgments

This work was supported by the Ministerio Español de Ciencia e Innovación under project number PID2020-115570GB-C22 MCIN/AEI/10.13039/501100011033 and by the Cátedra de Empresa Tecnología para las Personas (UGR-Fujitsu).